\documentclass[11pt]{article}

\usepackage[margin=1in]{geometry}
\usepackage{amsmath, amssymb}
\usepackage{booktabs}
\usepackage{graphicx}
\usepackage{caption}
\usepackage{subcaption}
\usepackage{hyperref}
\usepackage{natbib}
\usepackage{array}
\usepackage{enumitem}
\usepackage{xcolor}

\hypersetup{
  colorlinks=true,
  linkcolor=blue!50!black,
  citecolor=blue!50!black,
  urlcolor=blue!50!black
}

\usepackage{setspace}
\title{Pattern Selectivity is Not Task-Causal Structure:\\
A Cross-Architecture Mechanistic Study of Composed-Task Circuits\\
in 1B-Class Language Models\thanks{Correspondence: \texttt{abbyxu@gmail.com}. Code, data, and reproducibility scripts: \url{https://github.com/skydancerosel/spectral-probe-circuits}}}

\author{Yongzhong Xu}
\date{}

\begin{document}

\maketitle

\begin{abstract}
\noindent
We test whether the recipe of identifying attention-head circuits by task-pattern selectivity and causal ablation produces consistent mechanistic claims across model families. The recipe ports across pipelines; the specific circuit it identifies does not. Across four composed tasks (indirect-object identification, greater-than, successor sequences, variable binding) and three 1B-class language models from distinct training pipelines (Pythia 1B / Pile / dense; OLMo 1B / DCLM / dense; OLMoE 1B-7B / DCLM / MoE), we run a unified screen-and-ablate protocol with a matched-random null sampled across ten seeds per cell. The resulting 12 (task, model) cells have no two that share the same primary causal screen at comparable effect size. We introduce a five-category screen-outcome taxonomy --- \textbf{primary cause}, \textbf{secondary cause}, \textbf{correlate}, \textbf{interferer}, \textbf{null} --- with quantitative thresholds, and we show that all five outcomes appear in the panel. We propose a falsifiable hypothesis: the MoE model in our panel builds composed-task circuits on top of a foundational prev-token positional substrate (the prev-token-circuit ablation is the strongest causal screen on 3 of 4 tasks for OLMoE 1B-7B), with the IOI exception consistent with task-level final-position name-copying. We frame the methodology honestly: the spectral participation-ratio integral introduced in Paper~1 of this series is a \emph{general} indicator of specialized computation; what makes a finding task-specific is the task-pattern screen plus a per-model causal verification. This paper documents twelve such verifications across the 1B-class frontier and proposes a model-class-level mechanistic hypothesis tied to MoE architecture.
\end{abstract}

\section{Introduction}
\label{sec:intro}

The classical demonstration of mechanistic interpretability \citep{wang2022ioi} identified an attention-head circuit for indirect-object identification (IOI) in GPT-2 small: name-mover heads at late layers, S-Inhibition heads suppressing the subject, prev-token and induction heads in middle layers, duplicate-token and negative name-mover heads as auxiliary components. The circuit decomposition is canonical. The question this paper takes up is what happens when we apply the same conceptual decomposition --- the same family of attention-pattern \emph{types} --- to other 1B-class language models trained on different corpora with different architectures.

GPT-2 small does not solve IOI at 124M parameters (top-1 13\%, IO-vs-subject 57\%; see \S\ref{sec:ioi-baseline}). The smallest natural-text models in our panel that reliably solve IOI are at the 1B-active-parameter scale: Pythia 1B \citep{biderman2023pythia} trained on the Pile, OLMo 1B \citep{groeneveld2024olmo} trained on DCLM, and OLMoE 1B-7B \citep{muennighoff2024olmoe} trained on DCLM with a mixture-of-experts architecture (64 experts, top-8). These three models span two architecture families (dense, MoE) and two training corpora (Pile, DCLM), giving a small but real cross-pipeline panel at the scale where IOI is solvable.

We ask three operational questions:
\begin{enumerate}[topsep=2pt,itemsep=2pt]
\item Does a single task-pattern screen identify the IOI circuit in all three 1B-class models? --- \emph{No.} Four candidate screens (prev-token, induction, name-mover, S-Inhibition) cover IOI across the panel, but the \emph{primary} screen differs by model: prev-token-primary in Pythia, S-Inhibition-primary in OLMo, name-mover-primary in OLMoE.
\item Is the IOI cross-architecture decoupling a quirk of IOI, or a general feature of how 1B-class models implement composed tasks? --- We test three additional composed tasks: greater-than \citep{hanna2023greater}, successor sequences \citep{gould2023successor}, and a variable-binding task. The 4-task $\times$ 3-model grid (12 cells) contains no two cells with the same primary causal screen at the same effect size.
\item Does any cross-model structural pattern persist? --- \emph{Yes, one.} OLMoE 1B-7B's primary causal screen on 3 of 4 tasks (greater-than, successor, variable binding) is the prev-token circuit rather than the task-specific screen identified directly from the task structure. The fourth task (IOI) is name-mover-primary, consistent with IOI being a final-position name-copying task whose structure directly probes a different attention pattern. We call this the \emph{OLMoE prev-token-primacy} pattern and state it as a falsifiable cross-MoE hypothesis.
\end{enumerate}

\textbf{Contributions.}
\begin{itemize}[topsep=2pt,itemsep=2pt]
\item A 4-task $\times$ 3-model empirical grid (12 cells) of unified screen-and-ablate analyses at the 1B-class frontier, with a matched-random control sampled across ten seeds per cell to give per-cell specificity differentials with uncertainty (\S\ref{sec:matched-random}).
\item A five-category screen-outcome taxonomy --- primary cause, secondary cause, correlate, interferer, null --- with quantitative thresholds (\S\ref{sec:taxonomy}). All five categories appear in the panel.
\item A falsifiable cross-model mechanistic hypothesis (\S\ref{sec:olmoe-hypothesis}): MoE models at 1B-active scale build composed-task circuits on top of a foundational prev-token positional substrate, with exceptions only when the task structure directly probes a different attention pattern. Predictions for other MoE language models (Mixtral, DBRX, OLMoE-7B-A1.7B) follow.
\item A methodological refinement of the matched-random control (\S\ref{sec:matched-random}): for screens whose heads concentrate in early layers (L0--L1, where input-embedding processing is critical), same-layer matched-random has high variance and a weak null floor; we report when the bound applies and when it does not.
\item Honest framing of the recipe: the spectral PR-integral is a general specialization indicator, the task-pattern screen is what makes a finding task-specific, and individual-head ablations are the diagnostic for distinguishing supporters from interferers when group ablations alone are ambiguous.
\end{itemize}

\textbf{Companion papers.} This is Paper~3 of a three-paper series. Paper~1 \citep{paper1_methodology} introduces the screen-and-ablate recipe and validates it on capability-class tasks (induction, prev-token). Paper~2 \citep{paper2_developmental} characterizes the formation timeline of these circuits across pretraining. The present paper takes the recipe as given and asks how it behaves at the composed-task frontier, across architectures.

\section{Related Work}
\label{sec:related}

\textbf{Composed-task circuits in small models.} \citet{wang2022ioi} introduced the IOI task and its head-class decomposition in GPT-2 small. \citet{hanna2023greater} characterized GPT-2 small's greater-than circuit. \citet{gould2023successor} identified ``successor heads'' --- attention heads whose OV circuit increments along ordinal sequences --- in GPT-2 small, Pythia 410M, and several other models, finding the same head class recurs. These three works form the empirical basis for our task-specific screens.

\textbf{Induction heads and capability circuits.} \citet{olsson2022induction} characterized induction heads as attention heads implementing $A\ B\ \ldots\ A \to B$ copy patterns and connected their formation to in-context-learning emergence. \citet{elhage2021mathematical} laid out the framework for analyzing transformer circuits at the attention-head level.

\textbf{Automated circuit discovery.} \citet{conmy2023acdc} developed ACDC, an iterative edge-pruning algorithm for circuit identification. \citet{marks2024sparsefeature} extended automated discovery using sparse autoencoders to identify monosemantic features across models. Anthropic's monosemanticity work \citep{templeton2024monosemanticity} demonstrated feature-level interpretability scaling. Our recipe is complementary: head-level granularity, screen-and-ablate per task pattern, no model retraining required.

\textbf{Cross-architecture circuit transfer.} \citet{lieberum2023chinchilla} examined whether circuit-level findings on small models port to Chinchilla, finding that the same task uses different specific heads. \citet{marks2024sparsefeature} document partial feature-level overlap across model families using SAE dictionaries. The IOI cross-architecture result in the present panel sharpens this: not only do specific heads differ across models, the \emph{type} of attention pattern that carries the causal signal differs.

\textbf{Attention sinks.} \citet{xiao2024streaming} introduced the ``attention sink'' phenomenon: pretrained language models reliably allocate large attention probability to the first token regardless of content. The methodology paper of this series documents that BOS-class heads (best-class first-token attention sink) make up 43\%--78\% of all attention heads in 1B-class models, scaling with training data (DCLM $>$ Pile) and architecture (dense $>$ MoE). This paper takes that finding as background context: variable-binding screens in BOS-dominated regimes pick up BOS-class confound heads as their top candidates, and the diagnostic for separating real circuits from BOS-confounds is the individual head ablation (\S\ref{sec:vb}).

\textbf{The 1B-class frontier.} The Pythia, OLMo, and OLMoE pretrained model suites at the 1B-active-parameter scale represent the first scale at which IOI and other composed tasks become reliably solvable in our panel. GPT-2 124M does not solve IOI. The mechanistic-interpretability literature has predominantly studied either much smaller models (GPT-2 124M, Pythia 410M) or much larger ones (Chinchilla, Claude 3 Sonnet); the 1B-class frontier where multiple training pipelines coexist at the same scale has not been systematically characterized at the composed-task level across architectures. The present work is a small step in that direction.

\section{Methodology Recap}
\label{sec:methods}

We summarize the screen-and-ablate recipe from Paper~1 \citep{paper1_methodology}. The three steps are:

\begin{enumerate}[topsep=2pt,itemsep=2pt]
\item \textbf{Spectral signal.} For each (layer, head) and training checkpoint, compute the participation ratio of the per-head attention-output singular-value distribution over a fixed evaluation batch:
\[
\mathrm{PR}(L,H,t) = \exp\!\big( -\textstyle\sum_i p_i \log p_i \big), \quad p_i = \sigma_i^2 / \textstyle\sum_j \sigma_j^2.
\]
The trajectory feature $I(L,H) = \sum_t \max(\mathrm{PR}_t - 1, 0)\,\Delta\log(\text{tokens}_t)$ weights sustained content-dependent computation and surfaces heads doing specialized work.
\item \textbf{Task-pattern screen.} For each candidate attention pattern, compute a per-head selectivity (the head's mean attention into the pattern's target position relative to a uniform-other baseline). Filter for selectivity above a fixed threshold; this produces a small (typically 3--13-head) candidate circuit.
\item \textbf{Causal verification.} Group-ablate the candidate circuit (mean-ablation: replace each head's output with the batch-mean activation) and report $\Delta$top-1 and $\Delta$logit-diff vs. baseline. The control: matched-random heads in the same layers, no overlap with the picks, equal head count.
\end{enumerate}

\textbf{Honest framing.} The spectral PR-integral is a \emph{general} indicator of specialized computation. In attention-sink-dominated 1B-class models it does not by itself isolate the task-specific heads --- the top-K by PR-integral is dominated by generic content-dependent L0/L1 heads. What makes a finding task-specific is the task-pattern screen plus the causal verification.

\textbf{IOI-specific screens.} Four screens cover the Wang et al.\ IOI head-class decomposition:
\begin{itemize}[topsep=2pt,itemsep=2pt]
\item \textbf{Prev-token.} Best-class is previous-token at $\geq 100\times$ selectivity.
\item \textbf{Induction.} Best-class is induction at $\geq 50\times$ selectivity.
\item \textbf{Name-mover.} At the final query position, attention to the indirect-object name position relative to subject-name positions:
\[ \mathrm{nm\_sel} = \frac{\text{mean\_attn}(q \to \text{IO})}{\max(\text{mean\_attn}(q \to \text{subj1}),\, \text{mean\_attn}(q \to \text{subj2}))}. \]
\item \textbf{S-Inhibition.} At the final query position, attention to subject positions:
\[ \mathrm{subj\_sel} = \frac{\max(\text{subj1}, \text{subj2})}{\max(\text{io\_attn}, \varepsilon)}. \]
Filter to $\mathrm{subj\_sel} \geq 2$ and $\mathrm{subj\_max} \geq 0.1$, rank by $\mathrm{subj\_max}$.
\end{itemize}

\textbf{Matched-random control structure.} For each (task, model, screen) cell where the screen produces a non-trivial effect, we draw 10 independent random subsets of $k$ heads in the same layers as the screen picks, with no overlap with the picks. The resulting null is reported as mean $\pm$ std across the 10 seeds. The specificity differential is $|\Delta_\text{screen}| / |\Delta_\text{matched-random}|$ when both are sizable, with the caveat that for screens whose picks concentrate in L0--L1 (input-embedding-processing layers), the same-layer null has high variance.

\section{Setup}
\label{sec:setup}

\subsection{Three 1B-class models}

\begin{table}[h]
\centering
\small
\resizebox{\textwidth}{!}{%
\begin{tabular}{lllll}
\toprule
Model & Active params & Architecture & Training data & Tokenizer family \\
\midrule
Pythia 1B & 1B & GPT-NeoX, dense & The Pile & GPT-NeoX \\
OLMo 1B-0724-hf & 1B & Llama-style, dense & DCLM-aligned & OLMo \\
OLMoE 1B-7B-0924 & 1B / 7B total & Llama-style, MoE (64 experts, top-8) & DCLM & OLMoE \\
\bottomrule
\end{tabular}}
\caption{The three 1B-class models. Each entry differs from each other along at least one of architecture / training data / tokenizer; OLMo and OLMoE share the DCLM data and Llama-style architecture but differ in MoE.}
\label{tab:models}
\end{table}

\subsection{Cross-tokenizer batches}

All four task batches are constructed so that every token in every prompt is single-token in all three tokenizers (Pythia GPT-NeoX, OLMo, OLMoE). This eliminates tokenization confounds in cross-model comparisons.

\textbf{IOI batch.} 500 prompts, 50/50 ABBA/BABA mix, 42 single-token names, 6 places, 6 objects, RNG seed 42. Prompt: \emph{``When \{name1\} and \{name2\} went to the \{place\}, \{subject\} gave a \{object\} to ''} $\to$ predict the indirect object.

\textbf{Greater-than batch.} 500 prompts, RNG seed 42. Template: \emph{``The \{noun\} lasted from the year \{Y1\} to the year \{CC\}''} $\to$ model completes with a 2-digit token; correct iff the completion is greater than the start decade. 23 single-token nouns, 6 centuries (14--18; 19xx years BPE-merge in all three tokenizers and are excluded), 87 decades.

\textbf{Successor batch.} 118 unique 5-item ordinal sequences across four sequence types: days (cyclic 7), months (cyclic 12), ordinals (10, no wrap), numbers 1--99 (no wrap). Each item is single-token across all three tokenizers; prompts prepended with a leading space to use leading-space token forms.

\textbf{Variable binding batch.} 500 prompts, RNG seed 42. Template: \emph{`` \{name\_A\} lives in \{city\_A\}. \{name\_B\} lives in \{city\_B\}. \{query\_name\} lives in''} $\to$ predict the city bound to query\_name. 13-token deterministic prompt; 42 single-token names, 30 single-token cities, 50/50 query split.

\subsection{Evaluation}

For each task, at the final (target) position we measure top-1 accuracy (does the argmax equal the correct answer?), $\mathrm{logit\_diff} = \mathrm{logit}(\text{correct}) - \mathrm{logit}(\text{distractor})$, and $\mathrm{frac}(\text{correct} > \text{distractor})$ where both correct and distractor are well-defined for the task.

\section{IOI across three architectures}
\label{sec:ioi}

\subsection{Baseline capability}
\label{sec:ioi-baseline}

\begin{table}[h]
\centering
\small
\begin{tabular}{lrrr}
\toprule
Model & top-1 & frac(IO $>$ subj) & logit\_diff \\
\midrule
Pythia 1B & 90.4\% & 99.8\% & +4.00 \\
OLMo 1B & 87.8\% & 100.0\% & +3.64 \\
OLMoE 1B-7B & 58.6\% & 99.6\% & +3.95 \\
\midrule
GPT-2 124M & 13\% & 57\% & --- \\
\bottomrule
\end{tabular}
\caption{IOI baseline capability. All three 1B models solve IOI in the sense that the IO logit exceeds the subject logit on $\sim$100\% of prompts. OLMoE's lower top-1 reflects non-name tokens (function words, punctuation) outranking both names on $\sim$41\% of prompts; IO/subject preference is just as strong as in the dense models. GPT-2 124M does not solve IOI at this scale, so this is the first capability in our panel where small models cannot serve as sanity checks.}
\label{tab:ioi-baseline}
\end{table}

\subsection{Four screens $\times$ three models}
\label{sec:ioi-four-screens}

\begin{table}[h]
\centering
\small
\begin{tabular}{lrrrrr}
\toprule
Model & baseline & prev-token & induction & name-mover & S-Inhibition \\
\midrule
Pythia 1B & 90.4\% & \textbf{$\Delta$-82} & $\Delta$-1 & $\Delta$+7 (correlate) & $\Delta$-33 (5$\times$, secondary) \\
OLMo 1B & 87.8\% & $\Delta$+6 & $\Delta$-3 & $\Delta$-17 (1.5$\times$, partial) & \textbf{$\Delta$-32 ($\infty$, primary)} \\
OLMoE 1B-7B & 58.6\% & $\Delta$+26 & $\Delta$-1 & \textbf{$\Delta$-18 (30$\times$, primary)} & $\Delta$+0.6 (logit\_diff -3.0) \\
\bottomrule
\end{tabular}
\caption{IOI $\Delta$top-1 (percentage points) under each of the four screens, per model. Each row's primary screen is bolded. Same task, four candidate screens, three models, three different primary screens; no two cells are the same.}
\label{tab:ioi-grid}
\end{table}

The capability-screen ablations (prev-token, induction) use the standard thresholds from Paper~1: prev-token best-class at $\geq 100\times$ selectivity, induction at $\geq 50\times$. The IOI-specific screens (name-mover, S-Inhibition) use the definitions in \S\ref{sec:methods}. The numbers shown are the screen-ablation $\Delta$top-1 (percentage points) against the corresponding matched-random control reported in the next subsection.

\subsection{Three mechanistic implementations of the same task}
\label{sec:ioi-three-impls}

Reading across Table~\ref{tab:ioi-grid}, each model has a different primary screen, and each implementation has interpretive content beyond the methodology point.

\paragraph{Pythia 1B (Pile, dense): prev-token primary, with a redundant late-layer pathway.}
Ablating the prev-token circuit destroys IOI ($\Delta$-82pp). The name-mover-pattern heads are \emph{correlates} of IOI rather than causes --- ablating them alone helps IOI by 7pp, and the entire late-layer set helps by 8pp. By the time the residual stream reaches the layers where attention-to-IO is strong (L9+), the answer is already encoded by upstream prev-token computation, and the late-layer attention pattern reads off that encoding rather than constructing it.

S-Inhibition has a real \emph{secondary} causal role in Pythia: top-5 ablation drops top-1 by 33pp against a matched-random null of 6pp (5$\times$ differential). The substantive finding is the \textbf{name-mover + S-Inhibition union ablation}: top-1 drops to 37.6\% (a 53pp drop) and \emph{logit\_diff flips sign from +4.00 to -0.47}. The model now actively prefers the subject. The union of two screens, neither of which is primary on its own, suffices to invert the model's preference. Two observations:
\begin{itemize}[topsep=2pt,itemsep=2pt]
\item The two pathways operate at different layers (prev-token at L1--L7, name-mover/S-Inhibition at L8--L13). Pythia's IOI implementation has at least one early-layer route and one late-layer route, and the late-layer route alone --- with the early-layer route intact --- can flip the model's logit\_diff sign when ablated as a unit. This is a mechanistic finding about Pythia's IOI implementation, not just a methodology point.
\item The union ablation does not just degrade IOI; it inverts the model's preference. This is evidence of a \emph{functionally redundant} IOI pathway: the prev-token-dominated mechanism is the headline, but the late-layer name-mover + S-Inhibition pathway is also load-bearing and capable of driving the prediction by itself when combined.
\end{itemize}

\paragraph{OLMo 1B (DCLM, dense): S-Inhibition primary, with an infinite differential.}
Top-5 S-Inhibition ablation drops top-1 by 32pp; matched-random in the same layers actually \emph{helps} by 5pp. The differential is essentially infinite --- not just a large effect, but a control that goes the opposite direction. This is the cleanest primary-screen identification in the panel.

The name-mover screen partially captures OLMo's circuit ($\Delta$-17 vs.\ -11 matched-random, 1.5$\times$ differential --- real but distributed). The prev-token screen is wrong for IOI in OLMo entirely ($\Delta$+6). OLMo's IOI circuit lives in subject-attending late-layer heads (L8, L10, L11, L13). The mechanism is classical S-Inhibition: heads at the query position read the subject token and write a suppressive signal into the residual stream that reduces the subject's output logit.

\paragraph{OLMoE 1B-7B (DCLM, MoE): name-mover for the argmax, S-Inhibition for the margin.}
Top-5 name-movers ablation drops top-1 by 18pp; matched-random in the same layers drops it by 0.6pp. The differential is 30$\times$ (refined to 34$\times$ in the n=10 sweep, \S\ref{sec:matched-random}), by far the cleanest screen for this model. S-Inhibition has \emph{no top-1 effect} ($\Delta$+0.6, smaller than the matched-random $\Delta$-1.6) but produces a substantial logit\_diff drop (+3.95 $\to$ +0.95). Ablating S-Inhibition heads in OLMoE keeps the IO as argmax on nearly every prompt but collapses the margin by which the IO outranks the subject.

This is a distinction with real interpretive content. The IO-vs-subject competition in OLMoE has at least two computational components:
\begin{itemize}[topsep=2pt,itemsep=2pt]
\item A \emph{winner} component, driven by the name-mover circuit, that determines which token has the highest logit at the final position.
\item A \emph{margin} component, driven by S-Inhibition, that determines by how much the winner outranks the loser.
\end{itemize}
These are separable in OLMoE in a way they are not in Pythia or OLMo. Future work measuring IOI on OLMoE-style models should report both top-1 and logit\_diff; reporting only top-1 would miss the S-Inhibition contribution entirely.

\subsection{Top-5 candidates per screen}

\begin{table}[h]
\centering
\small
\begin{tabular}{p{2.5cm}p{11cm}}
\toprule
Model & Top-5 name-mover candidates (L\_H, nm\_sel) \\
\midrule
Pythia 1B & L13\_H1 (19.93), L10\_H6 (9.09), L10\_H5 (7.25), L10\_H4 (5.66), L11\_H7 (4.66) \\
OLMo 1B & L12\_H9 (4.25), L12\_H8 (3.75), L12\_H4 (2.78), L14\_H1 (2.33), L13\_H10 (2.25) \\
OLMoE 1B-7B & L12\_H14 (7.66), L12\_H3 (4.88), L12\_H10 (2.33), L11\_H8 (2.20), L9\_H1 (2.17) \\
\midrule
Model & Top-5 S-Inhibition candidates (L\_H, subj\_max, subj/io) \\
\midrule
Pythia 1B & L12\_H1 (0.50, 3.7$\times$), L9\_H1 (0.39, 5.1$\times$), L8\_H7 (0.32, 13.0$\times$), L12\_H4 (0.26, 6.2$\times$), L9\_H6 (0.20, 6.4$\times$) \\
OLMo 1B & L13\_H3 (0.47, 6.5$\times$), L11\_H5 (0.29, 9.8$\times$), L11\_H8 (0.24, 2.4$\times$), L12\_H1 (0.19, 5.0$\times$), L10\_H15 (0.17, 2.5$\times$) \\
OLMoE 1B-7B & L13\_H2 (0.55, 8.1$\times$), L8\_H2 (0.45, 19.0$\times$), L13\_H5 (0.43, 6.0$\times$), L13\_H1 (0.32, 4.7$\times$), L11\_H13 (0.26, 3.5$\times$) \\
\bottomrule
\end{tabular}
\caption{Top-5 candidates for the two IOI-specific screens across the three 1B models. Pythia has one dominant name-mover (L13\_H1, nm\_sel $\approx 20$ --- much sharper than any head in OLMo or OLMoE, where top nm\_sel is 4--8 and the signal is distributed). OLMo's L12\_H8 and OLMoE's L12\_H14 are also in their respective induction circuits --- multi-role late-layer heads.}
\label{tab:ioi-candidates}
\end{table}

\section{The matched-random n=10 specificity sweep}
\label{sec:matched-random}

A skeptical reading of the multi-screen procedure would be that trying many screens until one produces a large ablation effect is a fishing expedition. The matched-random differential --- same layers, equal head count, no overlap with the picks --- addresses this directly. For every (task, model, screen) combination where the screen produces a top-1 effect, we re-ran the matched-random control with 10 random seeds. The screen-specific result is from the original seed-123 run; the matched-random column is the n=10 mean $\pm$ std.

\begin{table}[h]
\centering
\small
\resizebox{\textwidth}{!}{%
\begin{tabular}{llrrr}
\toprule
Task & Screen & Pythia 1B & OLMo 1B & OLMoE 1B-7B \\
\midrule
IOI & top-5 name-mover & +7 / +2.7 $\pm$ 4.0 (correlate) & -17 / -9.6 $\pm$ 8.8 (1.8$\times$) & \textbf{-18 / -0.5 $\pm$ 5.1 (34$\times$)} \\
IOI & top-5 S-Inhibition & \textbf{-33 / -4.4 $\pm$ 7.3 (7.4$\times$)} & \textbf{-32 / -6.6 $\pm$ 9.3 (4.8$\times$)} & +1 / -10.6 $\pm$ 13.5 (none) \\
Greater-than & top-5 GT-specific & \textbf{-69 / +0.0 $\pm$ 0.0 ($\infty$)} & -1 / +0.0 $\pm$ 0.0 (margin) & -5 / -0.0 $\pm$ 0.1 (small) \\
Successor & top-5 self-attention & -38 / -21.0 $\pm$ 29.5 (1.8$\times$) & -81 / -57.5 $\pm$ 18.1 (1.4$\times$; L0) & -2 / -21.3 $\pm$ 31.7 (none) \\
\bottomrule
\end{tabular}}
\caption{The matched-random n=10 specificity sweep. Each cell reports $\Delta$top-1 (percentage points) for the screen / n=10 matched-random mean $\pm$ std. The ``X$\times$'' is the ratio of $|\Delta_\text{screen}|$ to $|\Delta_\text{matched-random}|$ when both are sizable. ``L0'' marks L0-concentrated screens where the same-layer null has high variance.}
\label{tab:mr-sweep}
\end{table}

\textbf{What the n=10 sweep adds beyond single-seed.} Three refinements.

\paragraph{1. The OLMoE $\times$ IOI-name-mover specificity holds dramatically.} The single-seed comparison gave 30$\times$; the n=10 sweep gives matched-random mean $-0.5 \pm 5.1$pp, refining the specificity to \textbf{34$\times$}. This is the largest screen-vs-random differential in the panel, and it survives the stronger control.

\paragraph{2. L0-concentrated screens have large matched-random variance, requiring care.} OLMo $\times$ Successor: matched-random mean $-57.5$pp with std $18.1$pp; the screen at $-80.5$pp is $1.4\times$ the random mean --- real, but a 23pp gap on a $>$50pp baseline, not the 2$\times$ that the single-seed comparison suggested. Pythia $\times$ Successor: matched-random mean $-21.0 \pm 29.5$pp (range $[-73, +1]$); the screen at $-38$pp is $1.8\times$ the random mean. \emph{L0 heads do critical input processing in these models}; removing any 5 of them is destructive regardless of which 5, so the specificity bound is weaker than for screens whose heads spread across many layers.

\paragraph{3. For Pythia $\times$ Greater-than, the matched-random std is essentially zero ($0.0 \pm 0.0$pp), confirming an effectively infinite specificity ratio for the GT-specific screen.} Of the 12 cells, the GT cases have the cleanest specificity differentials because the screen heads are distributed across 5 layers (L4, L6, L7, L8, L11) --- ablating any 5 random heads in those non-L0 layers does almost nothing.

\textbf{Methodological consequence.} The single most important consequence: \textbf{L0-concentrated screens cannot use same-layer matched-random as a tight null.} For OLMo and OLMoE successor in particular, the prev-token-circuit ablation (which is not L0-concentrated and has its own clean matched-random control) is the more reliable causal claim than the L0-concentrated successor screen.

The differentials still cluster into four categories --- weakly specific, strongly specific, saturated, or null --- but with the n=10 sweep we report mean and std rather than a single comparison. The ``screen is finding something specific'' claim survives this stronger control in every cell where the original single-seed comparison was strong; in cells where the single-seed comparison looked moderate (e.g., OLMo $\times$ IOI-name-mover at 1.5$\times$), the n=10 specificity is similar (1.8$\times$) --- the conclusion stands but the ratio should not be over-interpreted.

\section{Greater-than across three architectures}
\label{sec:gt}

\subsection{Setup and baseline}

The greater-than task \citep{hanna2023greater}: the model completes \emph{``The \{noun\} lasted from the year \{Y1\} to the year \{CC\}\_\_\_''} where $Y1 = CC\,DD$ with $DD \in [02, 88]$, and is correct iff the completion is a 2-digit token $> DD$.

\begin{table}[h]
\centering
\small
\begin{tabular}{lrrr}
\toprule
Model & top-1 & frac(year $>$ start) & logit\_diff (above - below) \\
\midrule
Pythia 1B & 99.6\% & 97--98\% & +4.3 \\
OLMo 1B & 99.6\% & 97--98\% & +5.0 \\
OLMoE 1B-7B & 99.6\% & 97--98\% & +4.8 \\
\bottomrule
\end{tabular}
\caption{Greater-than baseline. All three 1B models solve greater-than essentially perfectly.}
\label{tab:gt-baseline}
\end{table}

\subsection{GT-specific screen and top-5 candidates}

The task-pattern screen is attention from the final query position (\texttt{pos 11}) to the start-decade position (\texttt{pos 7}) in each prompt; high selectivity = candidate greater-than head.

\begin{table}[h]
\centering
\small
\begin{tabular}{ll}
\toprule
Model & Top-5 GT candidates (by attn(query $\to$ start-decade) / mean-other) \\
\midrule
Pythia 1B & L8\_H5, L7\_H0, L6\_H7, L11\_H6, L4\_H1 \\
OLMo 1B & L12\_H8, L2\_H7, L2\_H11, L4\_H12, L8\_H1 \\
OLMoE 1B-7B & L10\_H5, L10\_H2, L5\_H10, L12\_H10, L7\_H0 \\
\bottomrule
\end{tabular}
\caption{Top-5 greater-than candidates per model.}
\label{tab:gt-candidates}
\end{table}

\subsection{Ablation: three different profiles}

\begin{figure}[h]
\centering
\includegraphics[width=\textwidth]{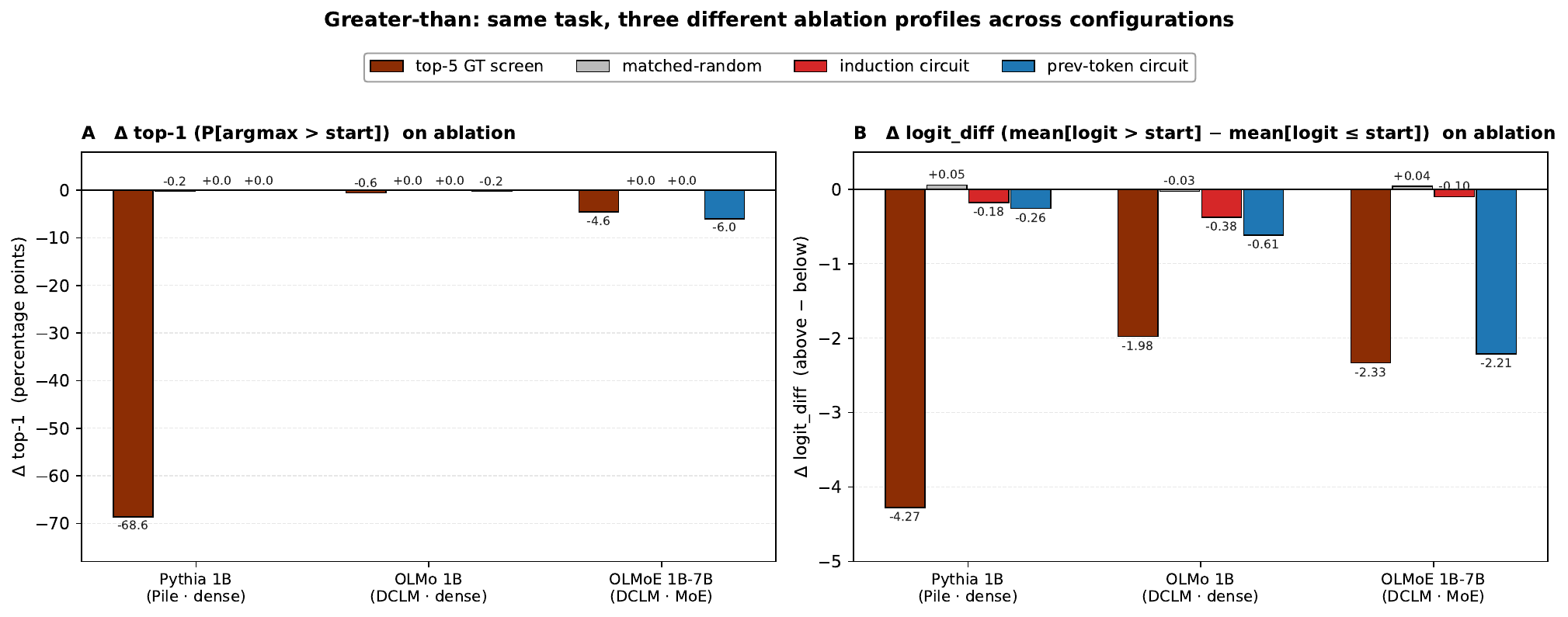}
\caption{\textbf{Greater-than: same task, three different ablation profiles across configurations.} (A) $\Delta$top-1 ($P[\text{argmax is a 2-digit number} > \text{start}]$, percentage points) on each ablation condition vs.\ baseline, for the three 1B-class configurations. (B) $\Delta$logit\_diff (mean[logit over above] - mean[logit over below]) on the same conditions. Conditions: \emph{top-5 GT screen} (heads selected by attention from query position to start-decade position; task-specific), \emph{matched-random} (5 random heads in the same layers as the top-5 GT picks, no overlap; null control), \emph{induction circuit} (heads with induction selectivity $\geq 50\times$ on the synthetic induction batch; a different screen on a different task), \emph{prev-token circuit} (heads with best-class prev-token and selectivity $\geq 100\times$). Three different ablation profiles emerge: Pythia 1B's GT is concentrated in the 5 GT-specific heads ($\Delta$top-1 = -68.6pp; everything else $<$ 1pp); OLMo 1B is top-1-robust to every ablation tried but the GT-screen compresses the logit margin ($\Delta$logit\_diff = -1.98 with $\Delta$top-1 only -0.6); OLMoE 1B-7B's prev-token circuit ablation hurts greater-than more than the GT-specific screen does ($\Delta$top-1 = -6.0 vs.\ -4.6; both above the matched-random null of 0).}
\label{fig:gt}
\end{figure}

\subsection{Three findings beyond the methodology point}

\paragraph{1. The Pythia 1B GT-specific heads are heterogeneous on the standard 6-class capability screen.}
Of the 5 GT-specific heads, one (L7\_H0) is a canonical induction head (induction-sel 113$\times$), one (L4\_H1) is a strong BOS attention-sink with multi-role classification (first-token 1433$\times$, prev-token 73$\times$, self 70$\times$, local 39$\times$), two (L8\_H5, L6\_H7) are weak first-token heads, and one (L11\_H6) is unclassified at the $30\times$ threshold. The GT screen selects them not because they share a single capability class but because they all route attention from \texttt{pos 11} to \texttt{pos 7} on this prompt structure. Crucially, the strongest Pythia induction head L4\_H4 is \emph{not} in the GT top-5 --- it does induction-pattern routing on the synthetic induction batch (where the routing target is ``after the duplicate'') but not on the greater-than prompt structure (where the target is ``decade of the first year''). ``Attention pattern'' and ``task-causal role'' are dissociable in Pythia 1B not because pattern-heads are not causal heads, but because the \emph{same head} can be both, neither, or one-but-not-the-other depending on prompt structure.

\paragraph{2. The margin-not-argmax signature appears in two (model, task) pairs.}
In OLMoE 1B-7B on IOI (\S\ref{sec:ioi-three-impls}), S-Inhibition ablation shifted logit\_diff from +3.95 to +0.95 without changing top-1. In OLMo 1B on greater-than, the top-5 GT-screen ablation shifts logit\_diff by -1.98 with top-1 changing by only -0.6pp. Two data points, two different (model, task) pairs, same signature: ablating the screen-identified circuit compresses the output-distribution margin while leaving the argmax robust. Top-1 accuracy is the wrong primary metric for ablation studies in these models; logit-margin or distributional measures are needed to see the effect. The mechanistic implication: some capabilities in larger / more redundant models are implemented as biases on the output distribution rather than as gating decisions about the argmax --- a distributed-redundant architecture in which ablating the primary heads weakens but does not remove the capability.

\paragraph{3. OLMoE 1B-7B builds greater-than on top of the prev-token circuit.}
Ablating OLMoE's prev-token circuit (8 heads identified by the standard prev-token-best-class screen at $\geq 100\times$ selectivity) hurts greater-than top-1 by 6.0pp (logit\_diff -2.21), \emph{more} than ablating the top-5 GT-specific heads identified by the GT screen ($\Delta$top-1 -4.6, $\Delta$logit\_diff -2.33). The same prev-token circuit ablation hurts top-1 by 0 in Pythia 1B and by 0.2pp in OLMo 1B, so the effect is OLMoE-specific. The compositional reading: OLMoE's greater-than computation is built on top of a positional substrate (the prev-token mechanism for routing back to the start-year position) rather than on a dedicated GT mechanism. The GT-specific heads do something secondary --- perhaps the comparison itself or the final readout --- but the heavy lifting is upstream in the prev-token circuit. This is one data point; the cross-MoE hypothesis follows in \S\ref{sec:olmoe-hypothesis}.

\section{Successor sequences across three architectures}
\label{sec:succ}

\subsection{Setup and baseline}

The successor task \citep{gould2023successor}: 5-item ordinal sequences across four sequence types (days, months, ordinals, numbers 1--99). The model predicts the next item.

\begin{table}[h]
\centering
\small
\begin{tabular}{lrrr}
\toprule
Model & top-1 accuracy & target\_logit\_mean & target\_rank\_median \\
\midrule
Pythia 1B & 78.8\% & +12.6 & 0 \\
OLMo 1B & 80.5\% & +13.7 & 0 \\
OLMoE 1B-7B & 76.3\% & +11.5 & 0 \\
\bottomrule
\end{tabular}
\caption{Successor baseline. All three solve the task --- median target rank = 0 on most prompts.}
\label{tab:succ-baseline}
\end{table}

\subsection{The ``successor head'' attention signature is itself model-dependent}

Per \citet{gould2023successor}, successor heads attend strongly to the current token (the most-recent shown item) and apply an OV-circuit transformation that increments to the next-in-sequence. The screen: per-head attention from \texttt{pos T-1} to itself, normalized to a uniform-other baseline.

\begin{table}[h]
\centering
\small
\begin{tabular}{ll}
\toprule
Model & Top-5 successor candidates (self-attention at \texttt{pos T-1}) \\
\midrule
Pythia 1B & L3\_H5 (succ\_sel 9.5, attn\_self 0.07, attn\_prev 0.91), L2\_H1, L8\_H5, L1\_H4, L7\_H0 \\
OLMo 1B & L0\_H2 (succ\_sel 4431, attn\_self 0.98), L0\_H10, L0\_H1, L0\_H0, L0\_H13 (all L0) \\
OLMoE 1B-7B & L0\_H15 (succ\_sel 9688, attn\_self 0.95), L0\_H14, L10\_H5, L7\_H10, L1\_H4 \\
\bottomrule
\end{tabular}
\caption{Top-5 successor candidates per model. OLMo and OLMoE's top successor heads attend $\geq 78\%$ to self at the query position (Gould's classical successor pattern). Pythia 1B's top successor head L3\_H5 attends 91\% to \emph{prev} and only 7\% to self --- the screen selects it because the prev-attention concentrates probability mass away from ``other'' positions, but the mechanism is prev-token-like rather than self-attention-like. Pythia's successor mechanism is structured differently.}
\label{tab:succ-candidates}
\end{table}

\subsection{Ablation: three more profiles}

\begin{figure}[h]
\centering
\includegraphics[width=\textwidth]{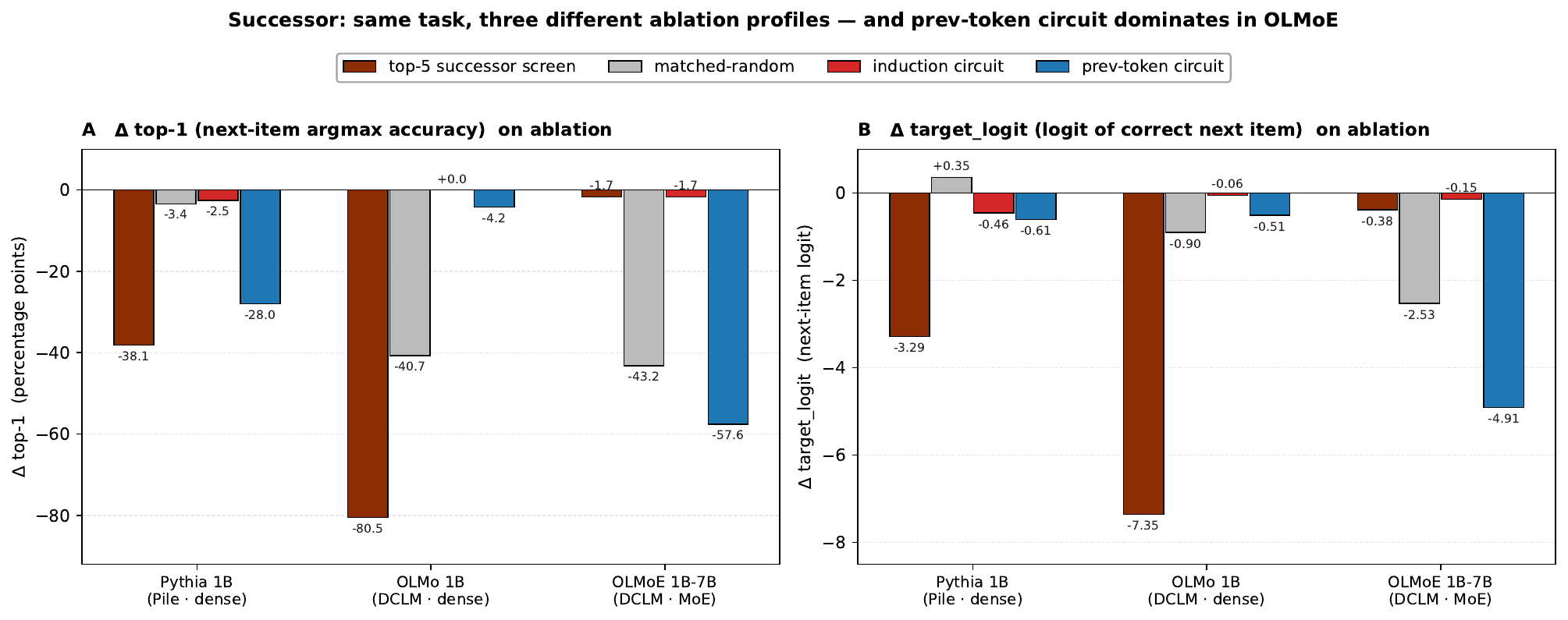}
\caption{\textbf{Successor sequences: same task, three different ablation profiles.} (A) $\Delta$top-1 ($P[\text{argmax} = \text{correct next item}]$, percentage points) under each ablation. (B) $\Delta$target\_logit (raw logit of the correct next item). Conditions match Figure~\ref{fig:gt} except \emph{top-5 successor screen} replaces \emph{top-5 GT screen}. Three profiles: \textbf{Pythia 1B} has mixed mechanism --- top-5 successor screen drops top-1 by 38.1pp \emph{and} prev-token circuit drops top-1 by 28.0pp (both above matched-random $\Delta$-3.4); \textbf{OLMo 1B}'s top-5 screen tanks top-1 by 80.5pp but matched-random in the same layers also drops it by 40.7pp because all 5 candidates are in L0 (where input-embedding processing happens); \textbf{OLMoE 1B-7B}'s prev-token circuit drops top-1 by 57.6pp --- far more than the top-5 successor screen's 1.7pp drop.}
\label{fig:succ}
\end{figure}

\subsection{The 3-task $\times$ 3-model grid (after successor)}

\begin{table}[h]
\centering
\small
\begin{tabular}{lp{3.5cm}p{3.5cm}p{4cm}}
\toprule
Task & Pythia 1B (Pile dense) & OLMo 1B (DCLM dense) & OLMoE 1B-7B (DCLM MoE) \\
\midrule
IOI & prev-token ($\Delta$-82) & S-Inhibition ($\Delta$-32) & name-mover ($\Delta$-18) \\
Greater-than & top-5 GT-specific ($\Delta$-69) & margin-not-argmax & prev-token ($\Delta$-6 $>$ GT-specific $\Delta$-5) \\
Successor & top-5 succ + prev-token ($\Delta$-38, $\Delta$-28) & top-5 succ-specific ($\Delta$-81; L0) & prev-token ($\Delta$-58 $\gg$ succ-specific $\Delta$-2) \\
\bottomrule
\end{tabular}
\caption{Three tasks, three models, nine cells --- no two cells use the same primary screen at the same effect size. The methodology recipe generalizes; the specific circuit does not.}
\label{tab:three-task-grid}
\end{table}

\subsection{Recurring sub-pattern: prev-token primacy in OLMoE on GT and successor}

OLMoE's column has prev-token as the primary screen on both greater-than ($\Delta$-6 vs.\ GT-specific $\Delta$-5) and successor ($\Delta$-58 vs.\ successor-specific $\Delta$-2). The L0-concentrated successor screen for OLMoE is unreliable as a causal claim (matched-random $\Delta$-43 in L0), but the prev-token circuit ablation (not L0-concentrated) has a clean matched-random control. The compositional-substrate hypothesis from \S\ref{sec:gt} now has two data points; the fourth task (variable binding, \S\ref{sec:vb}) adds a third and motivates the OLMoE prev-token-primacy claim of \S\ref{sec:olmoe-hypothesis}.

\section{Variable binding and the interferer screen-outcome category}
\label{sec:vb}

\subsection{Setup and baseline}

Variable binding: predict the city bound to query\_name. Prompt: \emph{`` \{name\_A\} lives in \{city\_A\}. \{name\_B\} lives in \{city\_B\}. \{query\_name\} lives in''}.

\begin{table}[h]
\centering
\small
\begin{tabular}{lrrr}
\toprule
Model & top-1 (correct city) & logit\_diff & frac(target $>$ distractor) \\
\midrule
Pythia 1B & 49.0\% & +0.39 & 73.2\% \\
OLMo 1B & 79.0\% & +1.54 & 91.2\% \\
OLMoE 1B-7B & 91.0\% & +1.15 & 91.4\% \\
\bottomrule
\end{tabular}
\caption{Variable binding baseline. Pythia's baseline is striking --- 49\% top-1 (essentially chance against the two-city choice) despite logit\_diff = +0.39 and frac(t$>$d) = 73\%. The model has a weak preference for the correct city but produces non-name tokens often enough that top-1 does not reflect it. OLMo and OLMoE solve VB cleanly.}
\label{tab:vb-baseline}
\end{table}

\subsection{Ablation: an interferer screen in Pythia, prev-token-primary in OLMo and OLMoE}

\begin{figure}[h]
\centering
\includegraphics[width=\textwidth]{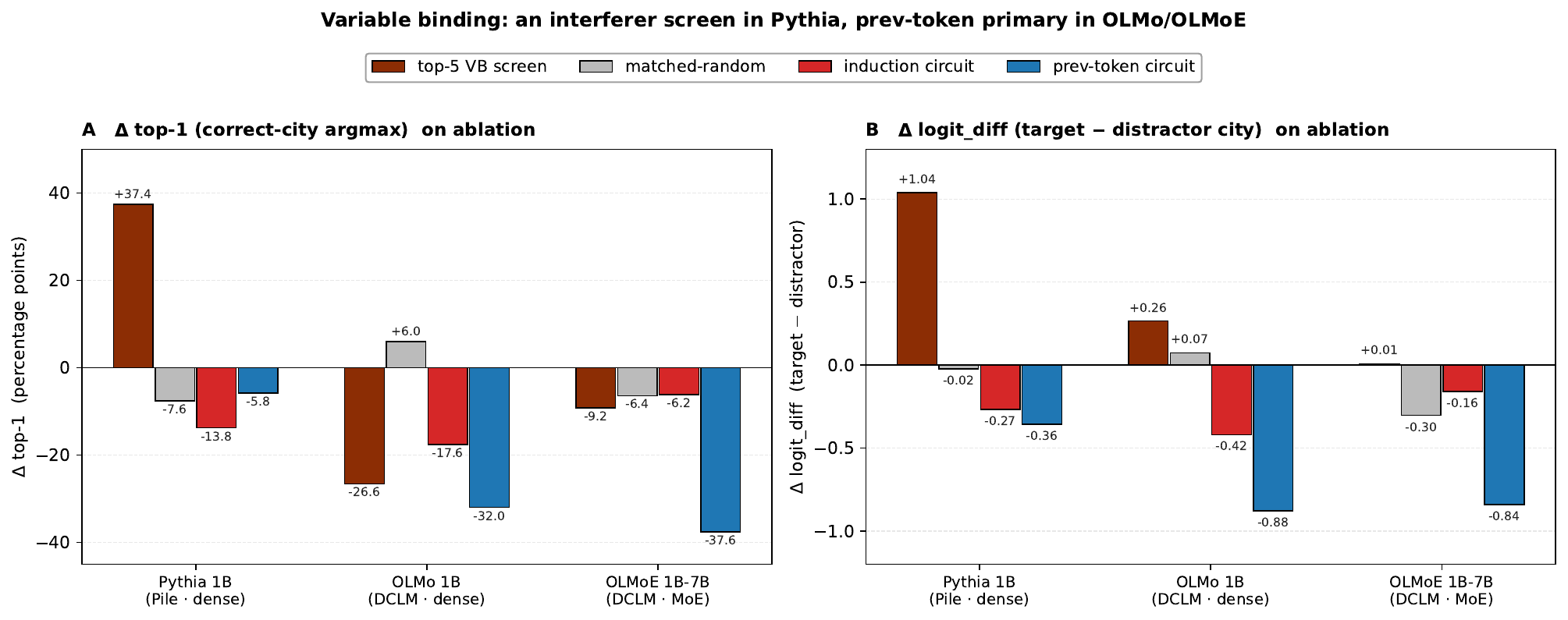}
\caption{\textbf{Variable binding: an interferer screen in Pythia, prev-token-primary in OLMo and OLMoE.} Same two-panel design as Figures~\ref{fig:gt} and \ref{fig:succ}. \textbf{Pythia 1B}: top-5 VB screen ablation \emph{increases} top-1 by +37.4pp (49\% $\to$ 86.4\%), logit\_diff rises by +1.04. The screen identifies heads that hurt the task. Matched-random in the same layers drops top-1 by 7.6pp; induction circuit -13.8pp; prev-token circuit -5.8pp. \textbf{OLMo 1B}: prev-token-circuit ablation hurts top-1 by 32.0pp; top-5 VB screen ablation hurts 26.6pp; induction circuit 17.6pp; the prev-token-circuit is primary, the VB screen is secondary, both above the matched-random null of +6.0pp. \textbf{OLMoE 1B-7B}: prev-token-circuit ablation hurts top-1 by 37.6pp; top-5 VB screen ablation hurts 9.2pp; induction and matched-random both around -6pp. The prev-token-circuit is overwhelmingly primary; the VB-specific screen is only weakly specific over the random baseline.}
\label{fig:vb}
\end{figure}

\subsection{The interferer category and the BOS-class diagnostic}

\paragraph{Pythia's naive VB-screen identifies interferer heads.}
The top-5 VB-screen heads in Pythia 1B (L8\_H5, L13\_H1, L7\_H2, L8\_H7, L12\_H1) all attend roughly equally to both the binding-value and the distractor positions (attn\_bind $\approx 0.30$--$0.40$, attn\_dist $\approx 0.20$--$0.34$, ratios 1.2--1.4$\times$). Their group ablation \emph{increases} top-1 by +37.4pp.

\paragraph{The cause is a BOS-class confound.}
4 of the 5 heads are best-classified as first-token (BOS) on the standard 6-class screen, with selectivities ranging from $30\times$ to $1112\times$. Their primary function is BOS attention-sink; their attention to the binding-value position is secondary. Ablating them removes their dominant BOS-attractor signal injection from the residual stream, which had been competing with the actual VB computation. Individual head ablations confirm: each of the 5 individually improves top-1 by +13 to +23pp (uniformly interferers; no heterogeneity).

\paragraph{The methodological fix: BOS-class filter (Pythia-specific).}
Re-filtering the VB-screen to exclude best-class = first-token heads: the top-5 non-BOS heads are L7\_H2 (induction), L4\_H4 (induction), L13\_H3 (self), L7\_H0 (induction), L0\_H2 (duplicate-token). Group ablation of this filtered set drops top-1 by $\boldsymbol{-16.2}$pp (49.0\% $\to$ 32.8\%) --- a real causal supporter circuit, recovered.

The ``interferer'' outcome on the naive screen was a confound from BOS-class heads sneaking into a non-BOS-related task-pattern screen via their secondary attention. With the BOS-class filter applied, Pythia VB's primary circuit is recoverable.

\paragraph{Cross-model test: the BOS-class filter does not generalize.}
Running the same diagnostic on the other two models reveals that the prescription is Pythia-specific:
\begin{itemize}[topsep=2pt,itemsep=2pt]
\item \textbf{OLMo 1B}: 5 of 5 top VB heads are best-class first-token (sel 59--228$\times$), but individual ablations show the set is heterogeneous: 3 supporters (L13\_H1 $\Delta$-6, L12\_H8 $\Delta$-10, L14\_H15 $\Delta$-17), 1 interferer (L12\_H9 $\Delta$+13), 1 null (L2\_H7 $\Delta$+0.4). The BOS-class-filtered top-5 (non-BOS heads: L11\_H13 self, L0\_H8 prev-token, L1\_H0 self, L12\_H15 self, L0\_H9 prev-token) ablates to $\Delta$-0.8 --- \emph{no causal effect}. In OLMo, the BOS-class heads \emph{are} the actual multi-role VB circuit; the BOS-class filter would remove the real circuit. (78\% of OLMo heads are BOS-class at the standard threshold; the model has so few non-BOS heads that the non-BOS pool simply does not contain the VB circuit.)
\item \textbf{OLMoE 1B-7B}: mixed best-class top-5 (3 first-token, 2 induction). Individual ablations all small ($-3$ to $+1.2$pp; group $-9$). The BOS-class-filtered top-5 (5 induction/self heads) ablates to $\Delta$+3 --- also a mild interferer. The actual OLMoE VB mechanism is the prev-token circuit ($\Delta$-37.6 established in Figure~\ref{fig:vb}), not anything the VB-screen picks up at top-5 regardless of filter.
\end{itemize}

\paragraph{General refined prescription: individual head ablations.}
The BOS-class filter is one useful tool but it does not generalize. The reliable methodological move is to run individual head ablations on every screen-identified circuit, especially in BOS-dominated regimes. Heads whose individual ablation hurts the task are \textbf{supporters}; heads whose individual ablation helps are \textbf{interferers}; heads with no individual effect are \textbf{nulls}. The group-ablation effect is approximately the sum of individual effects. The three model patterns underneath the same VB-screen output --- confound (Pythia), multi-role (OLMo), no-specific-circuit (OLMoE) --- illustrate that the screen output alone does not tell you which interpretation applies; the individual ablations do.

\section{The 4-task $\times$ 3-model grid and the screen-outcome taxonomy}
\label{sec:taxonomy}

\subsection{The full grid}

\begin{table}[h]
\centering
\small
\begin{tabular}{lp{3.5cm}p{3.5cm}p{4cm}}
\toprule
Task & Pythia 1B (Pile dense) & OLMo 1B (DCLM dense) & OLMoE 1B-7B (DCLM MoE) \\
\midrule
IOI & prev-token ($\Delta$-82) & S-Inhibition ($\Delta$-32) & name-mover ($\Delta$-18) \\
Greater-than & top-5 GT-specific ($\Delta$-69) & margin-not-argmax ($\Delta$logit -2.0) & prev-token ($\Delta$-6 $>$ GT-specific $\Delta$-5) \\
Successor & top-5 succ + prev-token ($\Delta$-38, $\Delta$-28) & top-5 succ-specific ($\Delta$-81; L0) & prev-token ($\Delta$-58 $\gg$ succ-specific $\Delta$-2) \\
Variable binding & VB-screen INTERFERER ($\Delta$+37); filtered $\Delta$-16; induction $\Delta$-14 & prev-token ($\Delta$-32) $>$ VB-specific ($\Delta$-27) & prev-token ($\Delta$-38 $\gg$ VB-specific $\Delta$-9) \\
\bottomrule
\end{tabular}
\caption{The 4-task $\times$ 3-model grid. Twelve cells, no two using the same primary screen at the same effect size. Each cell entry names the primary screen and its $\Delta$top-1. ``L0'' marks L0-concentrated screens whose matched-random null has high variance. The OLMoE column uses prev-token as primary on 3 of 4 tasks (GT, Successor, VB).}
\label{tab:full-grid}
\end{table}

\subsection{The five-category screen-outcome taxonomy}

Across the 12 cells, the screens produce qualitatively different kinds of outcomes. We name and quantify them:

\begin{enumerate}[topsep=2pt,itemsep=4pt]
\item \textbf{Primary cause.} Screen-$\Delta$ on the metric of interest is the largest non-saturated cell for the model, and the specificity differential is $\geq 3\times$ over matched-random.
\emph{Examples.} Pythia IOI prev-token ($\Delta$-82, $\infty$); Pythia GT top-5 GT-specific ($\Delta$-69, $\infty$); OLMo IOI S-Inhibition ($\Delta$-32, $\infty$); OLMoE IOI name-mover ($\Delta$-18, 34$\times$); OLMoE GT/Successor/VB prev-token.

\item \textbf{Secondary cause.} Screen-$\Delta$ is meaningful (typically $\geq 10$pp) and specificity differential is $\geq 3\times$, but a different screen has a larger primary effect on the same model+task. Often a redundant pathway.
\emph{Examples.} Pythia IOI S-Inhibition ($\Delta$-33, 7.4$\times$ --- name-mover + S-Inhibition union flips logit\_diff sign); OLMo VB top-5 VB-specific ($\Delta$-27 vs.\ prev-token $\Delta$-32); OLMo IOI name-mover ($\Delta$-17, 1.8$\times$, partial).

\item \textbf{Correlate.} Screen-$\Delta$ is small or zero, but the heads identified by the screen do show the expected attention pattern. Ablation of the late-layer set on which the screen lives shows the answer is already in the residual stream by that layer.
\emph{Example.} Pythia IOI name-mover ($\Delta$+7, helps; full late-layer set also helps by 8pp).

\item \textbf{Interferer.} Screen-$\Delta$ is positive (ablation \emph{helps} the task); individual ablations confirm the heads individually help the task when removed. Often a BOS-class confound (heads whose primary function is attention-sink, secondary role is the task-pattern attention).
\emph{Examples.} Pythia VB naive top-5 ($\Delta$+37); Pythia IOI name-mover ($\Delta$+7 with the full late-layer set $\Delta$+8 is a borderline correlate-vs-interferer case --- we class it as correlate because the heads do show the right attention pattern).

\item \textbf{Null.} Screen-$\Delta$ is within $\sim 1\sigma$ of matched-random ($\leq 2\times$ specificity). The screen identifies heads with the right attention pattern but no causal role in the task on this model.
\emph{Examples.} OLMoE IOI S-Inhibition (top-1 $\Delta$+1 vs.\ matched-random $\Delta$-11, no differential for top-1; logit\_diff differential exists separately and lifts this cell to ``margin-only secondary''); OLMoE IOI prev-token (helps by $\Delta$+26 but on no individual heads); OLMoE successor top-5 succ-specific ($\Delta$-2 vs.\ matched-random L0 $\Delta$-21, anti-specific).
\end{enumerate}

The empirical claim of the panel: \textbf{across 12 (task, model) cells, all five outcome categories appear, and no two cells share the same primary screen+effect-size profile.}

\subsection{Why this taxonomy matters}

Existing mechanistic interpretability work typically reports either ``circuit identified'' or ``no circuit''. The 12-cell panel demonstrates a richer space:
\begin{itemize}[topsep=2pt,itemsep=2pt]
\item A screen can identify heads that \emph{look} like the right circuit by attention pattern, yet have no causal role (correlate).
\item A screen can identify heads whose individual ablation \emph{helps} the task (interferer) --- and the cause can be a measurable BOS-confound that admits a diagnostic.
\item Multiple screens can identify partially-overlapping circuits on the same task on the same model, all with non-trivial causal roles (primary + secondary).
\item The same task on different models can use entirely different head classes as primary.
\end{itemize}
The taxonomy is what lets us report negative and positive results within the same framework, and what makes the matched-random differential interpretable.

\section{The OLMoE prev-token-primacy hypothesis}
\label{sec:olmoe-hypothesis}

\subsection{The pattern}

Across the four tasks in our panel:
\begin{itemize}[topsep=2pt,itemsep=2pt]
\item OLMoE 1B-7B's primary causal screen on greater-than is the prev-token circuit ($\Delta$-6 vs.\ GT-specific $\Delta$-5; \S\ref{sec:gt}).
\item On successor sequences, the prev-token circuit ablation drops top-1 by 57.6pp while the L0-concentrated successor screen drops it by 1.7pp; the prev-token circuit is overwhelmingly primary (\S\ref{sec:succ}).
\item On variable binding, the prev-token circuit ablation drops top-1 by 37.6pp while the VB-specific screen drops it by 9.2pp; again, prev-token primary (\S\ref{sec:vb}).
\item On IOI, OLMoE's primary screen is name-mover ($\Delta$-18, 34$\times$ specificity), and the prev-token circuit actively \emph{helps} when ablated ($\Delta$+26).
\end{itemize}

Three of four tasks: prev-token-primary in OLMoE. The exception is IOI. The dense models (Pythia, OLMo) show no such cross-task pattern --- their primary screens vary by task (prev-token / GT-specific / mixed for Pythia; S-Inhibition / margin-only / L0-specific for OLMo).

\subsection{The hypothesis}

\textbf{MoE models at 1B-active-parameter scale build composed-task circuits on top of a foundational prev-token positional substrate, except when the task structure directly probes a different attention pattern.}

The IOI exception is consistent: IOI is structured around final-position name-copying, an attention pattern directly probed by the name-mover screen (attention from the final query position to the IO name position). For IOI, the structural attention pattern \emph{is} the task; for GT, Successor, and VB, the structural attention pattern is upstream positional routing (prev-token) on top of which the task-specific computation happens.

A possible mechanistic explanation: MoE expert specialization requires a positional substrate to route through. With sparse activation patterns (top-8 of 64 experts per token, in OLMoE), the dense ``everywhere-to-everywhere'' connectivity of a fully dense transformer is replaced by sparse expert-mediated computation. The prev-token circuit is the natural positional backbone --- it provides a stable, attention-pattern-based way to route information from earlier positions to the current position, on top of which any task-specific computation can be built. Dense models trained on the same data (OLMo) have full dense attention plus a dense MLP, and can dedicate any subset of attention heads to a task-specific pattern; MoE models route the same computation through a sparser substrate where the prev-token circuit is load-bearing.

This is one of several possible mechanistic accounts. The hypothesis is stated to be falsifiable.

\subsection{Falsifiable predictions}

\begin{enumerate}[topsep=2pt,itemsep=2pt]
\item \textbf{Cross-MoE replication.} Other 1B-active-class MoE language models --- Mixtral 8x7B \citep{jiang2024mixtral}, DBRX \citep{databricks2024dbrx}, OLMoE-7B-A1.7B (if released) --- should show prev-token-primary on prev-token-compatible composed tasks (greater-than, successor, variable binding), and a task-structural screen primary on tasks like IOI whose structure directly probes a different pattern.
\item \textbf{Task-structural exception.} Within a single MoE model, tasks whose structure is final-position name-copying or final-position single-target-attention should have a non-prev-token primary screen, while tasks involving multi-position composition or positional routing should be prev-token primary.
\item \textbf{Dense-model contrast.} Dense models trained on the same data should \emph{not} systematically show prev-token-primacy on composed tasks. Our data (OLMo) confirm this: OLMo's primary screens across the four tasks are S-Inhibition, margin-only, succ-specific, prev-token-primary-on-VB (one out of four for OLMo, vs.\ three out of four for OLMoE).
\item \textbf{Architecture vs.\ data dissociation.} If a second MoE model were trained on Pile rather than DCLM (matching Pythia's data), the prev-token-primacy pattern should still appear. If a third dense model were trained on DCLM at 1B with the OLMo architecture, the lack of prev-token-primacy should still appear. These would jointly establish the pattern as architecture-driven rather than data-driven.
\end{enumerate}

\subsection{Limitations of the hypothesis}

The hypothesis is supported by three data points (GT, Successor, VB on OLMoE) plus one consistent exception (IOI). Three positives and one consistent exception is not a strong empirical case; it is a starting hypothesis with explicit predictions. The cross-MoE replication on Mixtral / DBRX is the necessary next step. The hypothesis would be falsified by an MoE model showing task-specific (non-prev-token) primary screens on multiple composed tasks at comparable scale.

\section{BOS-fraction context}
\label{sec:bos}

The 12-cell grid is shaped by a cross-panel invariant from Paper~1 of this series: the fraction of attention heads classified as BOS (first-token-attention-sink at $\geq 30\times$ selectivity) at the final checkpoint scales with training data and architecture.

\begin{table}[h]
\centering
\small
\begin{tabular}{lr}
\toprule
Model & BOS-class fraction \\
\midrule
Pythia 160M (Pile, dense) & 43.1\% \\
Pythia 410M (Pile, dense) & 58.1\% \\
Pythia 1B (Pile, dense) & 53.9\% \\
OLMoE 1B-7B (DCLM, MoE) & 68.0\% \\
OLMo 1B (DCLM, dense) & 78.1\% \\
\bottomrule
\end{tabular}
\caption{BOS-class fraction at the final checkpoint. Within Pythia, BOS fraction grows with scale (43\% $\to$ 58\%) and saturates near 54\% at 1B. DCLM data adds $\sim 20$pp over Pile at the same scale+architecture (OLMo 1B 78\% vs.\ Pythia 1B 54\%). MoE \emph{reduces} BOS by $\sim 10$pp vs.\ dense at the same scale+data (OLMoE 1B-7B 68\% vs.\ OLMo 1B 78\%). MoE does not cause attention sinks; if anything, it suppresses them relative to dense architecture trained on the same data.}
\label{tab:bos-frac}
\end{table}

\textbf{Why this matters for the present paper.} BOS-dominated regimes are where the screen-outcome taxonomy gets interesting. The interferer category, in particular, is BOS-confound-driven in Pythia VB: 4 of 5 top heads from the VB-screen are best-class first-token, and their individual ablation helps the task because they were primarily injecting BOS-attractor signal that competed with the VB computation. The BOS-class filter prescription works in Pythia precisely because Pythia's BOS fraction (54\%) leaves 46\% of heads as a sufficiently rich non-BOS pool to find the real circuit in. In OLMo (78\% BOS-class), the non-BOS pool is too small, and the BOS-class heads themselves are the real multi-role VB circuit; the filter would remove the actual circuit. The diagnostic that generalizes across all BOS regimes is the individual head ablation (\S\ref{sec:vb}), not the BOS-class filter itself.

\section{Discussion}
\label{sec:discussion}

\subsection{The recipe ports as a recipe; specific circuits do not}

Twelve (task, model) cells across two architecture families and two training pipelines. The screen-and-ablate procedure produces a meaningful causal verification in every cell --- a primary, secondary, correlate, interferer, or null outcome on each combination. The taxonomy is fine-grained enough to distinguish between mechanistic claims that are real causal supporters versus heads that merely show the right attention pattern.

But no two cells in the grid agree at the level of \emph{which specific screen is the primary causal one}. The IOI circuit on GPT-2 small (Wang et al., 2022) has name-movers as primary; the IOI circuit on Pythia 1B has prev-token as primary; the IOI circuit on OLMo 1B has S-Inhibition as primary; the IOI circuit on OLMoE 1B-7B has name-mover as primary (consistent with GPT-2-small, but in a different architecture family and at $\sim 8\times$ scale). The task is constant, the behavioral capability is constant ($\sim 100\%$ frac(IO $>$ subj) across all three 1B models), and the mechanistic implementation is not.

For an interpretability researcher applying the recipe to a new task on a new model: the recipe is the right starting point, but the specific circuit on a different model should be re-derived from scratch with the full family of candidate screens, not transferred from a known model. The recipe is a generative procedure; the specific circuit is a per-model empirical question.

\subsection{``What is the circuit for capability X'' is malformed for the 1B-class frontier}

The question ``what is the circuit for capability X'' presupposes a one-circuit answer. The 12-cell grid shows that the question, asked across model families, has no one-circuit answer at the 1B-class scale. There are convergent functional solutions across distinct training pipelines and architectures, all of which produce the same behavioral capability through different mechanistic implementations.

This is analogous to the situation in biology: ``what is the protein for vision'' has different answers in vertebrates, cephalopods, and arthropods --- all of which see, via different protein families and different evolutionary origins. The mechanistic-interpretability question becomes about the \emph{class} of solutions and how the class is structured, rather than the specific solution. The five-category taxonomy in \S\ref{sec:taxonomy} is one such structural classification; the OLMoE prev-token-primacy hypothesis in \S\ref{sec:olmoe-hypothesis} is a sub-class structural claim tied to MoE architecture.

\subsection{Top-1 accuracy is the wrong primary metric for ablation studies}

The margin-not-argmax signature appears in two (model, task) cells (OLMoE IOI under S-Inhibition; OLMo GT under top-5 GT-specific). In both, ablating the screen-identified circuit compresses the logit-margin while leaving top-1 essentially unchanged. Reporting only top-1 would miss the effect entirely. The mechanistic implication: some capabilities in larger / more redundant models are implemented as biases on the output distribution rather than as gating decisions about the argmax.

A distributed-redundant architecture has the property that the primary circuit's contribution is marginal-rather-than-categorical --- when ablated, the answer is still the argmax, but the model is less confident. The model-class implication: future cross-architecture circuit work should report $\Delta$top-1 \emph{and} $\Delta$logit-diff (and, ideally, full output-distribution divergences) as standard practice. The IOI cross-architecture literature on GPT-2 small could plausibly miss this margin-only effect if it exists at small scale --- a hypothesis that follows from our observation but is not tested in our panel.

\subsection{Implications for production-LLM circuit discovery}

If the cross-architecture mechanistic decoupling at 1B-class scale persists to the 70B+ frontier where current production LLMs live, then circuit discovery on, e.g., Llama 3 70B would not transfer to Mistral Large or Claude or GPT-4. Each frontier model would require its own family-of-screens analysis. This is consistent with the SAE-dictionary cross-architecture results of \citet{marks2024sparsefeature} and the model-by-model nature of the Anthropic monosemanticity work \citep{templeton2024monosemanticity}.

The OLMoE prev-token-primacy hypothesis sharpens this: there may be \emph{model-class-level} regularities (all 1B-class MoE models build composed-task circuits on a prev-token substrate) even when individual circuit specifics differ. The future of cross-model circuit discovery may be more about identifying the right model-class-level regularity and screening for it than about porting specific circuits.

\subsection{Companion-paper context}

The formation timeline of these circuits is the subject of a companion paper \citep{paper2_developmental}: across all three 1B-class configurations, the capability circuits (induction, prev-token) form within the first 0.3--2\% of training tokens, and the per-head spectral signal precedes their formation. The cross-task / cross-architecture mechanistic structure documented here is established at a single (final) checkpoint per model; whether the cross-architecture decoupling is also visible at intermediate checkpoints is an open question not addressed in the present work. The methodology paper \citep{paper1_methodology} introduces the recipe in its original form on capability-class tasks; the present paper takes the recipe to the composed-task frontier and extends it to a family-of-screens analysis.

\section{Limitations}
\label{sec:limitations}

\paragraph{Four tasks $\times$ three models is still a small grid.}
Twelve cells is enough to demonstrate that the 5-category taxonomy is non-empty (all categories appear) and to motivate the OLMoE prev-token-primacy hypothesis as a starting empirical claim, but not enough to establish strong cross-task generalization. Replication on more tasks (e.g., docstring completion, factual recall, in-context-learning span-copying) and more model families (Llama 1B, Qwen, MoE variants beyond OLMoE) would strengthen both the taxonomy and the hypothesis.

\paragraph{The interferer category rests on a single clear instance.}
Of the five outcome categories, \emph{interferer} is the thinnest-supported in the panel. The one unambiguous case is Pythia variable binding (\S\ref{sec:vb}), where the naive VB screen's top-5 are BOS-confound heads whose individual ablation \emph{helps} the task; the Pythia IOI name-mover is a borderline interferer-versus-correlate case that we class as correlate. So while all five categories appear, the interferer category is instantiated once with confidence. Whether it recurs across tasks and models, or is specific to BOS-confounded screens in moderate-BOS-fraction models like Pythia, is not established by this panel and would need additional cells to confirm.

\paragraph{Single seed per natural-text 1B model.}
The Pythia, OLMo, and OLMoE 1B models are each single pretrained checkpoints. Whether the \emph{primary screen choice} (prev-token for Pythia IOI, S-Inhibition for OLMo IOI, name-mover for OLMoE IOI) is seed-stable or seed-dependent is not testable without re-pretraining. The TS-51M six-seed experiment in Paper~1 shows that the \emph{specific induction heads} differ across pretraining seeds even on the same model and task. Whether the primary-screen \emph{type} is more stable than the specific heads, at the 1B scale, is an open empirical question.

\paragraph{OLMoE prev-token-primacy has 3 data points.}
The hypothesis rests on greater-than, successor, and variable binding on OLMoE 1B-7B, plus the consistent IOI exception. This is a small empirical base. Validation on Mixtral, DBRX, OLMoE-7B-A1.7B, or other 1B-active-class MoE models is the necessary next step. The hypothesis is stated in falsifiable form (\S\ref{sec:olmoe-hypothesis}) precisely so it can be tested by future cross-MoE work.

\paragraph{We do not measure why the screens disagree across models.}
The 12-cell grid documents that the primary screen differs; it does not characterize what residual-stream features carry the task-relevant signal in each model, or why one model uses one attention pattern when another uses another. This would require deeper interpretability work --- residual-stream feature decomposition, ideally via sparse autoencoders trained per model --- and is beyond the scope of the present screen-and-ablate study.

\paragraph{L0-concentrated screens have weak matched-random nulls.}
For OLMo successor (top-5 all in L0) and OLMoE successor (top-1 in L0), the same-layer matched-random control has high variance and a destructive null floor. The specificity differential is correspondingly weaker. For these cells we recommend the prev-token-circuit ablation (not L0-concentrated) as the more reliable causal claim. The methodology paper's matched-random control is a sound primary tool but needs the L0 caveat in future iterations.

\paragraph{Causal-target metrics restricted to top-1 and logit-diff.}
Both metrics are unidimensional. Per-example logit attribution, distribution-shift sensitivity (e.g., KL divergence on the output distribution), and granular per-prompt circuit dependence would all be informative. The margin-not-argmax signature highlights that logit-diff catches effects top-1 misses, but logit-diff itself is a one-number summary of a high-dimensional output distribution.

\section{Conclusion}
\label{sec:conclusion}

We applied a single screen-and-ablate recipe to 12 (composed task, 1B-class model) cells across three architectures and two training pipelines. The recipe works in every cell; the specific causal circuit it identifies differs in every cell. We introduced a five-category taxonomy --- primary cause, secondary cause, correlate, interferer, null --- and showed that all five appear in the panel. We proposed a falsifiable cross-MoE hypothesis tied to the prev-token positional substrate.

The methodology-level lesson is straightforward: the recipe ports, the specific circuit it identifies does not, and individual head ablations are the reliable diagnostic for separating supporters from interferers in BOS-dominated regimes. The model-class-level lesson is more provisional but more interesting: there appear to be MoE-specific structural regularities (the prev-token-primacy pattern) that hold across composed tasks within an architecture family. Whether this generalizes across MoE language models is the next empirical question.

\bibliography{refs}

\end{document}